\documentclass{article}
\usepackage{spconf,amsmath,graphicx}
\usepackage{bbm}
\usepackage{stfloats}
\usepackage{color,soul}
\usepackage{makecell}
\usepackage{float}
\usepackage{url}

\title{Fast, Accurate Barcode Detection in Ultra High-Resolution Images}
\name{Jerome Quenum, Kehan Wang, Avideh Zakhor\thanks{This work was supported by Amazon.com, Inc.}}

\address{
        Department of Electrical Engineering and Computer Science \\
        University of California, Berkeley\\
        \{jquenum, wang.kehan, avz\}@berkeley.edu}

\begin{document}
\ninept

\maketitle
\begin{abstract}
Object detection in Ultra High-Resolution (UHR) images has long been a challenging problem in computer vision due to the varying scales of the targeted objects. When it comes to barcode detection, resizing UHR input images to smaller sizes often leads to the loss of pertinent information, while processing them directly is highly inefficient and computationally expensive. In this paper, we propose using semantic segmentation to achieve a fast and accurate detection of barcodes of various scales in UHR images. Our pipeline involves a modified Region Proposal Network (RPN) on images of size greater than 10k$\times$10k and a newly proposed Y-Net segmentation network, followed by a post-processing workflow for fitting a bounding box around each segmented barcode mask. The end-to-end system has a latency of 16 milliseconds, which is $2.5\times$ faster than YOLOv4 and $5.9\times$ faster than Mask R-CNN. In terms of accuracy, our method outperforms YOLOv4 and Mask R-CNN  by a $mAP$ of 5.5\% and  47.1\% respectively, on a synthetic dataset. We have made available the generated synthetic barcode dataset and its code at \texttt{\url{http://www.github.com/viplabB/SBD/}}.
\end{abstract}

\begin{keywords}
Barcode detection with deep neural networks, barcode segmentation, Ultra High-Resolution images.
\end{keywords}

\section{Introduction}
\label{sec:intro}
Barcodes are digital signs often made of adjacent and alternating black and white smaller rectangles that have become an intrinsic part of human society. In administration, for example, they are used to encode, save, and retrieve various users' information. At grocery stores, they are used to track sales and inventories. More interestingly in e-commerce, they are used to track and speed up processing time in warehouses and fulfillment centers.

In classical signal processing, filters used for detection are image-specific since input images are not all necessarily acquired with the same illumination, brightness, angle, or camera. Consequently, adaptive image processing algorithms are required, which can impact detection accuracy \cite{Namane2017}. In addition, because classical signal processing methods often run on Central Processing Units, they tend to be much slower compared with deep learning implementations that are easily optimized on Graphics Processing Units (GPUs).

\begin{figure*}
  \includegraphics[width=\textwidth,height=6cm]{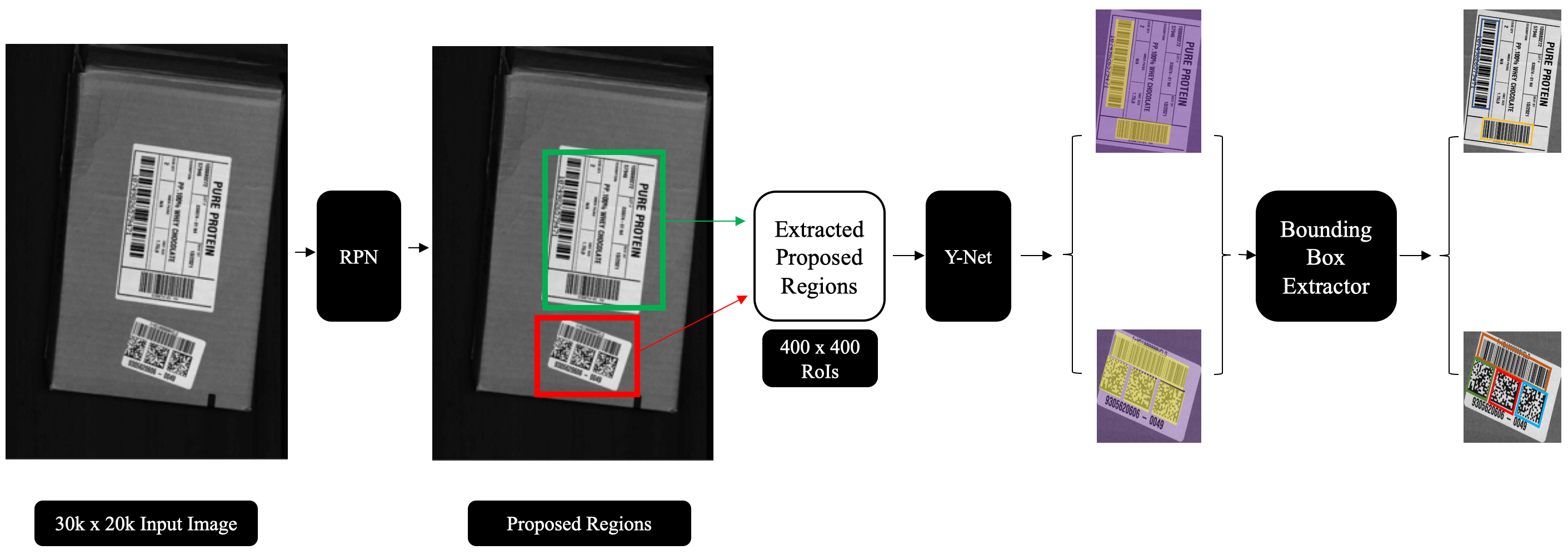}
  \caption{Proposed Approach, the modified RPN is followed by Y-Net and the bounding box extractor.}
  \label{ynet-full-pipeline}
\end{figure*}

Over the years, a number of methods have been proposed to detect barcodes using classical signal processing \cite{Namane2017,Hock2004,Katona2013,Soros2013,Creusot2016}, but nearly all of them take too long to process Ultra High-Resolution (UHR) images. More specifically, \cite{ Creusot2016} used parallel segment detectors which improved on their previous work \cite{Creusot20155} of finding imaginary perpendicular lines in Hough space with maximal stable extremal regions to detect barcodes. Katona et al. \cite{ Katona2013} used morphological manipulation for barcode detection, but this method did not generalize well as different barcode types have varying detection performances. Similarly, \cite{Gallo2011} proposed using x and y derivative differences, but varying input images yields different outputs, and using such operation on UHR images often become highly inefficient. 

With neural networks, though there has been much improvement in barcode detection tasks, few of them have addressed the fast and accurate detection problem in UHR images. Zamberletti et al. \cite{Zamberletti2010} paved the way for using neural networks to detect barcodes by investigating Hough spaces. This was followed by \cite{Hansen2017} which adapted the You Only Looked Once (YOLO) detector to find barcodes in Low Resolution (LR) images, but the YOLO algorithm is known to perform poorly with long shaped objects such as code 39 barcodes. Instance segmentation methods such as Mask R-CNN  \cite{He2017} perform better on images of size $1024\times1024$ pixels but on smaller size images, the outputted Region of Interests (RoI) do not align well with long, 1D-barcode structures. This is because it typically predicts masks on $28\times28$ pixels irrespective of object size, and thereby generates "wiggly" artifacts on some barcode predictions, losing spatial resolution. In the same way, dedicated object detection pipelines, such as YOLOv4 \cite{Bochkovskiy2020}, though they perform well on lower Intersection over Union (IoU) thresholds, suffer accuracy at higher IoU thresholds. Among those using segmentation on LR images as a means for detection, \cite{Zharkov2019} also tends to not perform well at higher IoU thresholds.

\begin{figure}[h] 
\centering
\includegraphics[height=3.1in]{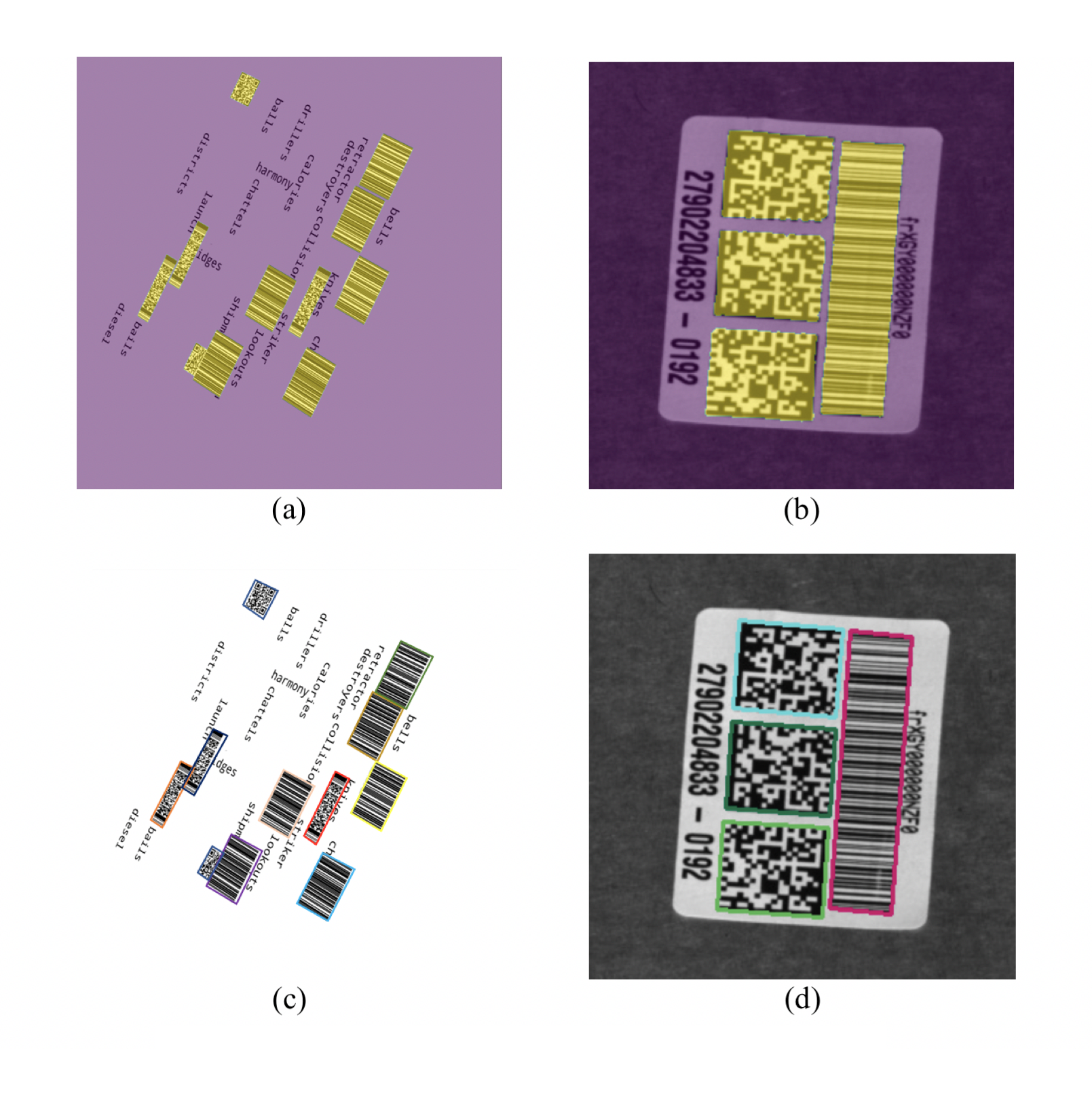}
\caption{ Sample outputs of our pipeline; yellow - segmented barcode pixels; purple - segmented background pixels; boxes - bounding box extracted; (a) synthetic barcode image; (b) real barcode image; (c) prediction results on (a); (d) prediction results on (b).}
\label{picture}
\end{figure}

In this paper, we propose a pipeline for detecting barcodes using deep neural networks, shown in {Fig. \ref{ynet-full-pipeline}}, which consists of two stages trained separately. When compared with classical signal processing methods, neural networks not only provide a faster inference time, but also yield higher accuracy because they learn meaningful filters for optimal feature extraction. As seen in {Fig. \ref{ynet-full-pipeline}}, in the first stage, we expand on the Region Proposal Network (RPN) introduced in Faster R-CNN \cite{Girshick2015} to extract high definition regions of potential locations where barcodes might be. This stage allows us to significantly reduce inference computation time that would have been required otherwise in the second stage. In the second stage, we introduce Y-Net, a semantic segmentation network that detects all instances of barcodes in a given outputted RoI image ($400\times400$). We then apply morphological operations on the predicted masks to separate and extract the corresponding bounding boxes as shown in {Fig. \ref{picture}}. 

One of the limitations of existing work on barcode detection is the insufficient number of training examples. ArTe-Lab 1D Medium Barcode Dataset \cite{Zamberletti2010} and the WWU Muenster Barcode Database \cite{Wachenfeld2010} are two examples of existing available datasets. They contain 365 and 595 images respectively, with ground truth masks at a resolution of $640\times480$. Most of the samples in the ArTe-Lab dataset have only one EAN13 barcode per sample image, and few of them in the Muenster database have more than one barcode instance on a given image. To address this dataset availability problem, we have released 100,000 UHR and 100,000 LR synthetic barcode datasets along with their corresponding bounding boxes ground truths, and their ground truth masks to facilitate further studies. The outline of this paper is as follows: in Section 2, we describe details of our approach; in Section 3, we summarize our experimental results and in Section 4, we conclude and expend on our future work. 

\section{Proposed Approach}
\label{sec:pagestyle}

As seen in {Fig. \ref{ynet-full-pipeline}}, our proposed method consists of three stages: the modified Region Proposal Network stage, our Y-Net \footnote{Our Y-Net architecture resembles the English alphabet letter “Y” and differs from \cite{mohammed2018ynet} which used a pre-trained encoder network that is augmented with an untrained mirrored network and a decoder network.} segmentation network stage, and the bounding box extraction stage.

\subsection{Modified Region Proposal Network}
\label{ssec:subhead}

Region proposals have been influential in computer vision and more so when it comes to object detection in UHR images. It is common in UHR images that barcodes are clustered in a small region of the image. To filter out most of the non-barcode backgrounds, we modified the RPN introduced in Faster R-CNN \cite{Girshick2015} to propose regions of barcodes for our next stages. By first transforming the UHR input image to an LR input image of size $256\times256$, the RPN was trained to identify blobs in LR images. Once a bounding box is placed around the identified blobs, the resulting proposed bounding box is remapped to the input UHR image by a perspective transformation, and the resulting regions are cropped out. The LR input to the RPN is chosen to be of size $256\times256$ as a lower resolution results in the loss of pertinent information. Non-Max Suppression (NMS) is used on the predictions to select the most probable regions.

\subsection{Y-Net Segmentation Network}
\label{ssec:subhead}

\begin{figure*}[h!] 
\centering
\includegraphics[width=17cm, height=10cm]{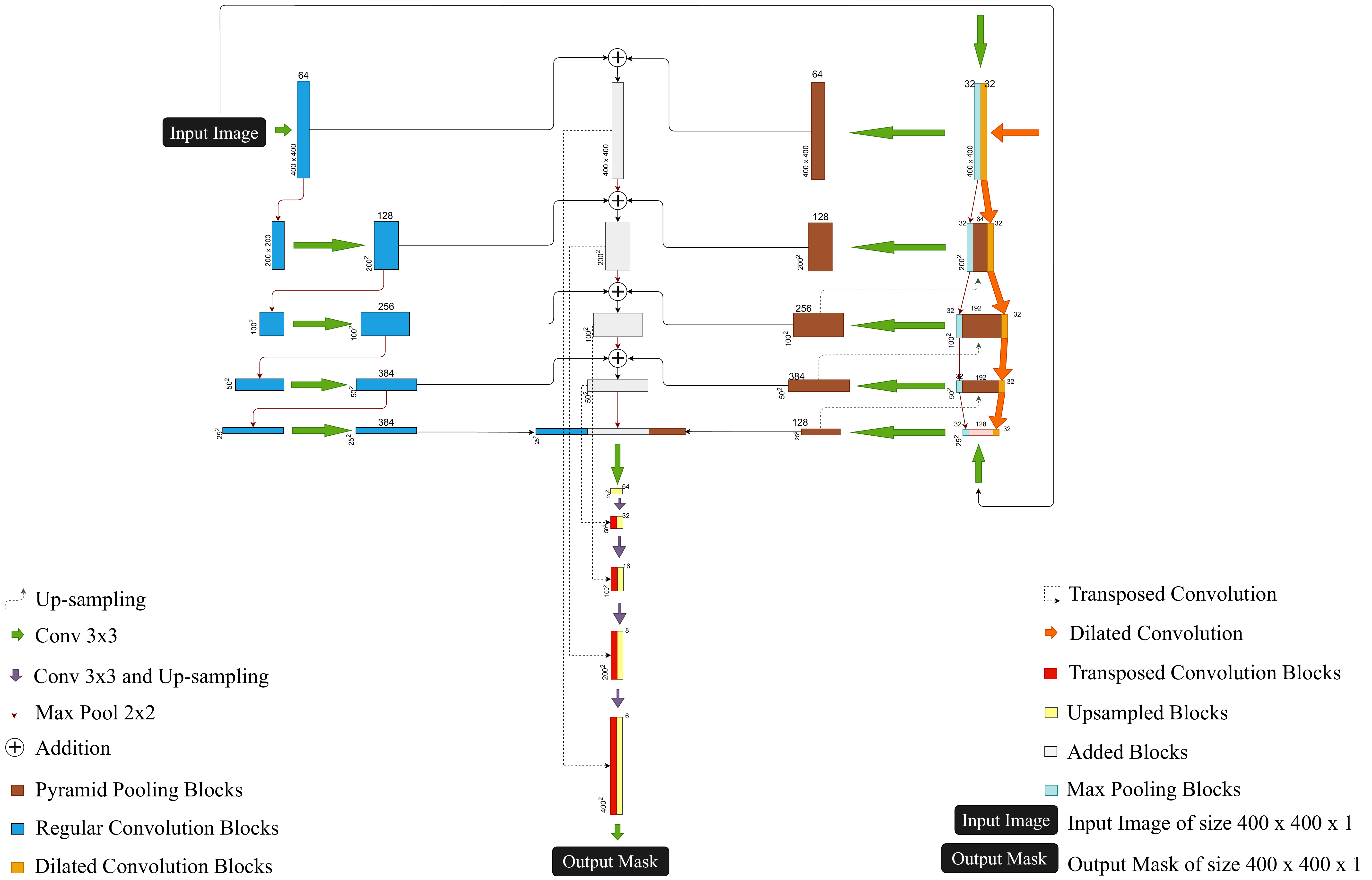}
\caption{ Y-Net Architecture.}
\label{ynet-picture}
\end{figure*}

As depicted in {Fig. \ref{ynet-picture}}, Y-Net is made out of 3 main modules distributed in 2 branches: a Regular Convolutional Module shown in blue which constitutes the left branch, and a Pyramid Pooling Module shown in brown, along with a Dilated Convolution Module shown in orange which after concatenation and convolution constitute the right branch. 

The \textbf{\textit{Regular Convolution Module}} takes in $400\times400$ output images of the RPN and consists of convolutional and pooling layers. It starts with 64 - channel $3\times3$ kernels and doubles the number at each layer. We alternate between convolution and max-pooling until we reach a feature map size of $25\times25$ pixels. This module allows the model to learn general pixel-wise information anywhere in the input image. 

The \textbf{\textit{Dilated Convolution Module}} takes advantage of the fact that barcodes have alternating black and white rectangles to learn sparse features in their structure. The motivation for this module comes from the fact that dilated convolution operators play a significant role in the \textit{"algorithme a trous"} for biorthogonal wavelet decomposition \cite{Holschneider1987}. Therefore, the discontinuities in alternating patterns and sharp edges in barcodes are more accurately learned by such filters. In addition, they leverage a multiresolution and multiscale decomposition as they allow the kernels to widen their receptive fields with dilation rates from 1 up to 16. Here too a $400\times400$ input image is used and we maintain 32 – channel $3\times3$ kernels throughout the module while the dimensions of the layers are gradually reduced using a stride of 2 until a feature map of $25\times25$ pixels is obtained.

The \textbf{\textit{Pyramid Pooling Module}} allows the model to learn global information about potential locations of the barcodes at different scales and its layers are concatenated with the layers on the dilated convolution module in order to preserve the features extracted from both modules. 

The resulting feature maps from the right branch are then added to the output of the Regular Convolution Module, which allows for the correction of features that would have been missed by either branch. In other words, the output of each branch constitutes a residual correction for the other thereby refining the result at each node as shown in white. The nodes are then up-sampled and concatenated with transposed convolution feature maps shown in red and yellow of the corresponding dimension. Throughout the network, we use ReLU as a non-linearity after each layer and add $L_2$ regularization to account for possible over-fitting scenarios that could have occurred during training. On all datasets, we use 80\% for the training set, 10\% for the validation set, and the remaining 10 \% for the testing set. We use one NVIDIA Tesla V100 GPU for the training process. Since this is a segmentation network and we are interested in classifying background and barcodes, we use binary cross-entropy as loss function.

\subsection{Bounding Box Extraction}
\label{ssec:subhead}

 Since some images contain barcodes that are really close to each other, their Y-Net outputs reflect the same configuration which makes the extraction of individual barcode bounding boxes complex as shown in {Fig. \ref{picture4}}(a). To separate them effectively,  we perform an erosion, contour extraction, and bounding box expansion with a pixel correction margin. As shown in {Fig. \ref{picture4}}(b), the erosion stage allows the algorithm to widen gaps between segmented barcodes that may be separated by 1 or more pixels. The resulting mask is then used to infer individual barcode bounding boxes in the contour extraction stage in {Fig. \ref{picture4}}(c) through border following. A pixel correction margin is used to recover the original bounding boxes' dimensions during the expansion stage as shown in {Fig. \ref{picture4}}(d). This post-processing stage of our pipeline has an average processing time of 1.5 milliseconds (ms) because it is made of a set of Python matrix operations to efficiently extract bounding boxes from predicted masks.



\begin{table*}[!t]  
  \centering
  \small
  \begin{tabular}{m{0.14\textwidth}|m{0.05\textwidth}m{0.05\textwidth}m{0.05\textwidth}m{0.05\textwidth}m{0.07\textwidth}||m{0.05\textwidth}m{0.05\textwidth}m{0.05\textwidth}m{0.05\textwidth}||m{0.05\textwidth}m{0.075\textwidth}}
     \Xhline{2\arrayrulewidth}
          & {$mAP$ (all)} & {$AP_{50}$ (all)} & {$AP_{75}$ (all)} & {$mAP$ (small}) & {$mAP$ (medium)} & {$AR_{50}$ (all)}    & {$AR_{70}$ (all)}    & {$AR_{80}$ (all)} & {$AR_{90}$ (all)} &{Latency (ms)} & {Resolution (px)} \\  \hline 
    Mask R-CNN \cite{He2017}    & .466          & .985          & .317          & .340          & .489   & .990         & .740         & .279       & .023 &   94.8             &	$448\times448$ \\
    YOLOv4 \cite{Bochkovskiy2020}      & .882          & \textbf{.990} & .989          & .815          & .897 & \textbf{1.} & \textbf{1.} & .995       & .873 &   40.5            &	$320\times320$ \\
    \textbf{Ours} & \textbf{.937} & \textbf{.990} & \textbf{.990} & \textbf{.903} & \textbf{.945} & \textbf{1.} & \textbf{1.} & \textbf{1.} & \textbf{.972} &\bf 16.0           & $400\times400$\\ 
    \Xhline{2\arrayrulewidth}
  \end{tabular}
\caption{\label{table1}Average Precision for Max Detection of 100 and Average Recall for Max Detection of 10 computed using MS COCO API.}
\end{table*}
\begin{table*}[!t]  
  \centering
  \small
\begin{tabular}{c|cccc||cccc}
  \Xhline{2\arrayrulewidth}
  &  \multicolumn{4}{c||}{Muenster Dataset} &  
     \multicolumn{4}{c}{ArTe Lab Dataset}  \\ 
     \hline
    
        &{DR} & {Precision} & {Recall} & {mIoU} & {DR} & {Precision} & {Recall} & {mIoU}   \\ 
        \hline
         
Creusot et al. \cite{Creusot2016} &	.982 &	        -       &	    -       &	-       &	.989 &	        -   &	        -          &	-        \\
Hansen et al. \cite{Hansen2017} &      .991 &	        -       &	    -       &	.873   &      .926 &	        -   &	        -          &	.816    \\
Namane et al. \cite{Namane2017} &     .966 &	        -        &	    -        &	.882    &     .930 &	        -   &	        -        &	.860    \\
Zharkov et al. \cite{Zharkov2019} &    .980 &	    .777        &	.990        &	.842    &    .989 &	        .814 &	        .995        &	.819     \\
\bf ours        &\bf 1. &   \bf .984        &\bf 1.         &\bf.921     &    \bf 1.   &      \bf .974 &          \bf 1.      &\bf	.934        \\  
\Xhline{2\arrayrulewidth}
  \end{tabular}
\caption{\label{table2}Mean IoU (mIoU), Precison and Recall and Detection Rate (DR) at IoU threshold of 0.5 (Muenster and ArTe-Lab Dataset).}
\end{table*}
\begin{table}[ht!]  
  \centering
  \small
  \begin{tabular}{c|ccccc}
    \Xhline{2\arrayrulewidth}
          & Px Acc      & Px mIoU     & Px Prec     & Px Rec   \\  
          \hline
    Mask R-CNN \cite{He2017}    & .993          & .990          & .989          & .890    \\
    \textbf{Ours} & \textbf{1.} & \textbf{1.} & \textbf{.999} & \textbf{.999 } \\ 
    \Xhline{2\arrayrulewidth}
  \end{tabular}
\caption{\label{table3}Pixel-wise Metrics}
\end{table}


\section{Datasets and Results}
\label{sec:majhead}


For the synthetic dataset, we use treepoem \footnote{https://github.com/adamchainz/treepoem} and random-word \footnote{https://github.com/vaibhavsingh97/random-word} to generate UHR and LR barcode images. We use Code 39, Code 93, Code 128, UPC, EAN, PD417, ITF, Data Matrix, AZTEC, and QR among others. We model the number of barcodes in a given image using a Poisson process and a combination of perspective transforms is used to make the barcodes vary in shape and position from one image to the other. We have also added random black blobs at random locations on the original UHR and LR canvases. The real UHR barcodes dataset obtained from Amazon.com, Inc is made of 3.8 million UHR images of resolution up to $30k\times30k$ grayscale images and could not be released due to confidentiality reasons. Additionally, the Muenster and Artelab datasets are used with some data augmentation schemes for more samples.

For the RPN, we accumulated the number of bounding boxes inside the proposed regions and divided it by the total number of ground truth bounding boxes. Our implementation yields an accuracy of  98.03\%  on the synthetic dataset at 10 ms per image and 96.8\% on the real dataset at 13 ms per image while the baseline \cite{Girshick2015} yields the same accuracies and an average latency over 2.5 seconds (s) per image for both datasets.

For Y-Net, we use the Microsoft (MS) COCO API, and Pixel-wise metrics to evaluate against \cite{He2017, Bochkovskiy2020}. By default, the MS COCO API configuration evaluates on \textit{small, medium} and \textit{large} areas objects but in our application, the largest detected barcode area is \textit{medium}. Since Y-Net is a segmentation network and does not output confidence scores for each segmented barcode, we propose using \textit{pseudo scores}, the ratio of the total number of nonzero pixels in a predicted mask to the total number of nonzero pixels in the corresponding ground truth mask at the location of a given object. 

{Table \ref{table1}} shows $mAP$ and $mAR$ values of the models on the synthetic dataset. As seen, our pipeline outperforms \cite{He2017}, and \cite{Bochkovskiy2020} by a $mAP$ of 47.1\% and 5.5\% and $AP_{75}$ of 67.3\% and 0.1\% respectively. Also shown in {Table \ref{table1}}, is a $mAR_{90}$ improvement of 94.9\% and 9.9\% on \cite{He2017} and, \cite{Bochkovskiy2020} respectively which highlights that Y-Net continues to yield better $mAR$ results even at higher IoU thresholds. Both our approach and \cite{Bochkovskiy2020} achieve an $AR_{50}$ of 100\% and outperform \cite{He2017} by 1\%.  For \textit{small} area barcodes, Y-Net outperforms \cite{He2017} and \cite{Bochkovskiy2020} by a $mAP$ of 56.3\% and 8.8\% and for \textit{medium} area barcodes, Y-Net displays a $mAP$ increase of 45.6\% and 4.8\% on \cite{He2017} and \cite{Bochkovskiy2020} respectively. In addition, {Table \ref{table3}}  reveals that Y-Net a has much better semantic segmentation performance than \cite{He2017}.  {Table \ref{table1}} displays that Y-Net performs at least $2.5 \times$ faster than the fastest of models \cite{He2017} and, \cite{Bochkovskiy2020} on LR images. 

\begin{figure}[H] 
\centering
\includegraphics[width=3.45in, height=2.1in]{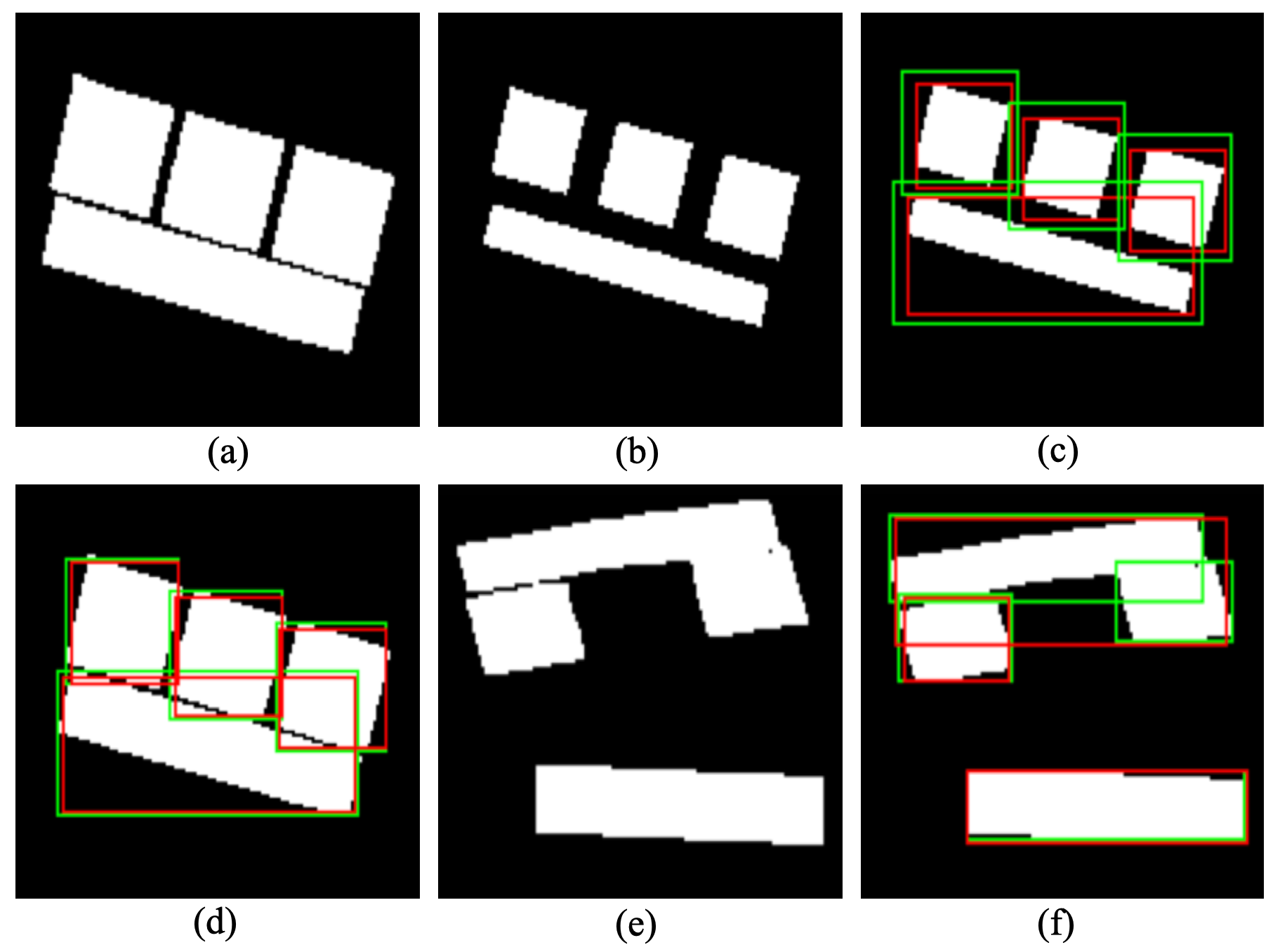}
\caption{ (a) Y-Net output; (b) Y-Net output after erosion; (c) extracted bounding boxes --red, ground truth bounding boxes --green on eroded output; (d) final bounding boxes after pixel correction margin on Y-Net output; (e) Y-Net output of occluded barcodes scenarios; (f) final extracted bounding boxes are grouped after pixel correction margin due to overlaping barcodes in input image.}
\label{picture4}
\end{figure}

Similarly, we have used the Detection Rate (DR), mIoU, Precision, and Recall, as described in \cite{Namane2017, Creusot2016, Hansen2017, Zharkov2019} on the Arte-Lab and Muenster datasets and as can be seen in {Table \ref{table2}}, our method outperforms previous works on all of the mentioned metrics. This indicates that our bounding box extraction algorithm is working as expected to detect accurate bounding boxes. 
However, while it is successful in separating barcodes that are relatively close to each other, it has limitations when barcodes are overlapping as shown in {Fig. \ref{picture4}}(e). For those occlusion scenarios, the algorithm tends to group the overlapping barcodes into one bounding box instead of separate bounding boxes as shown in {Fig. \ref{picture4}}(f).

\section{Conclusion}
\label{sec:print}



In this paper, we showed that barcodes can be efficiently, accurately, and speedily detected using Y-Net on UHR images. With \textit{pseudo scores} as \textit{confidence scores}, our approach outperforms existing detection pipelines with a much better latency. In future work, we aim to extend this method to the multi-class detection task for small objects in UHR images and videos in a weakly supervised fashion.

\clearpage
\bibliographystyle{IEEEbib}
\bibliography{strings,refs}
\end{document}